\documentclass{article}
\usepackage[utf8]{inputenc}
\usepackage{adjustbox}
\usepackage{algorithm}
\usepackage{algpseudocode}
\usepackage{booktabs}
\usepackage{amsmath, amssymb, mathtools}
\usepackage[english]{babel}
\usepackage{geometry}
\providecommand{\keywords}[1]{\textbf{\textit{Index terms---}} #1}
\title{gBeam-ACO: a greedy and faster variant of Beam-ACO}
\author{Jeff Hajewski\thanks{Work done while at The University of Iowa, Department of Computer Science}\\
Salesforce\\
jhajewski@salesforce.com
\and
Suely Oliveira\\
Department of Computer Science\\
University of Iowa
\and
David E. Stewart\\
Department of Mathematics\\
University of Iowa
\and
Laura Weiler\\
Department of Computer Science\\
University of Iowa}
\date{Preprint\\April 2020}

\begin{document}

\maketitle
\begin{abstract}
    Beam-ACO, a modification of the traditional Ant Colony
Optimization (ACO) algorithms that incorporates a modified beam search,
is one of the most effective ACO algorithms
for solving the Traveling Salesman Problem (TSP).
Although adding beam search to the ACO heuristic search process is effective, it
also increases the amount of work (in terms of partial paths) done by the
algorithm at each step. In this work, we introduce a greedy variant of Beam-ACO
that uses a greedy path selection heuristic.
The exploitation of the greedy path selection is offset by the
exploration required in maintaining the beam of paths. This approach has
the added benefit of avoiding costly calls to a random number generator and
reduces the algorithms internal state, making it simpler to parallelize.
Our experiments
demonstrate that not only is our greedy Beam-ACO (gBeam-ACO) faster than
traditional Beam-ACO, in some cases by an order of magnitude, but it does not
sacrifice quality of the found solution, especially on large TSP instances.
We
also found that our greedy algorithm, which we refer to as gBeam-ACO, was less
dependent on hyperparameter settings.
\end{abstract}

\section{Introduction}
Ant Colony Optimization (ACO) is an effective family of heuristic search
algorithms for the traveling salesman problem (TSP)
\cite{DBLP:books/sp/Reinelt94,gutin2006traveling}. The TSP is defined
as finding the shortest path in a graph $G=(V, E, d)$, where $d(x, y)$ is a
distance function,
\[
  d(x, y) : V\times V \rightarrow \mathbb{R}
\]
such that each vertex $v\in V$ is visited exactly once.
The variant of the TSP \cite{gutin2006traveling} we use in this paper is the
symmetric TSP in $\mathbb{R}^2$ that has the property
$d(x, y) = d(y, x)$  $\forall x, y \in V$.
We use the so-called \texttt{Euclidean2D} distance function, as defined in
TSPLIB \cite{tsplib,DBLP:journals/informs/Reinelt91}, where distance is calculated
as the rounded Euclidean distance between two points. That is
\[
  d(\mathbf{x}, \mathbf{y}) = \lfloor ||\mathbf{x} - \mathbf{y}||_2 + 0.5 \rfloor
\]
Additionally, we assume
a complete graph where each vertex is connected to every other vertex. Building
a valid path in this graph involves selecting a sequence of $|V|-1$
nodes from a starting node (referred to as the depot) where there is an implicit
connection from the last selected node to the depot. Finding the optimal
solution to the TSP requires evaluating all $|V|!$ paths.

Ant Colony Optimization algorithms construct this path by traversing the graph
using so-called ants and depositing so-called ``pheromone'' on the graph edges
based on the characteristics of the traversed paths (e.g., shorter paths result in more
pheromone). Pheromone levels on the edges of the graph encourage the ants to
explore parts of the graph that have previously resulted in shorter paths.

ACO has a number of variants that mainly differ in the pheromone update rule.
The original Ant System (AS) \cite{dorigo1992optimization} updated the pheromone
levels using the paths of all the ants in the colony.
Elitist\cite{DBLP:conf/fgit/JaradatA10,DBLP:journals/corr/abs-0811-0131} and 
Min-Max Ant System (MMAS) \cite{10.5555/348599.348603} are two of the most popular variants.
The Elitist ACO variant is characterized by only updating the pheoromone levels
with the shortest path traversed by the ant colony. MMAS, on the other hand,
is similar to AS in that each ant's path contributes to the pheromone update,
but also scales pheromone levels such that they are guaranteed to be between
user-specified minimum and maximum levels.

One particularly effective variant of ACO is Beam-ACO
\cite{blum2005beam,lopez2010beam,blum2008beam,caldeira2007supply,thiruvady2009hybridizing}
, which
adapts Beam search \cite{DBLP:journals/ai/MedressCFGKONNRRSWW77,zhang1998complete}
to the ACO algorithm.
As with other ACO implementations, Beam-ACO relies on a
pseudo-random number generator (PRNG). PRNGs can be very performant, such as the
PCG family of PRNGs \cite{oneill:pcg2014}; however, they are frequently a bottleneck in stochastic
algorithms. For example, prior work on particle swarm optimization noted that
the way in which the random numbers were generated for use by the algorithm had a material impact
on performance \cite{DBLP:conf/cec/HajewskiO19}. Stochastic algorithms require additional thought when
parallelizing because the PRNG may represent shared state, which can cause
contention when called from multiple threads. This is can be overcome
using techniques such as using thread-local PRNG functions or a
producer-consumer approach with a thread-safe queue, but none of these are
simpler than simply avoiding the need for a PRNG.

Greedy search heuristics are known to find sub-optimal solutions on many
problems, one of which is the Traveling Salesman Problem. The typical greedy algorithm
for the TSP is to select the shortest path to an unvisited vertex from the
current vertex.
This algorithm will select
the same path each iteration without any exploration.
We use the term \emph{iteration} to mean one generation of a path or paths for all
ants (in the case of ACO), followed by updating the pheromone matrix using an
 appropriate update rule.

Ant Colony Optimization algorithms use a heuristic function that combines edge
length and pheromone level.
The pheromone amounts change each iteration of the algorithm based on pheromone
``evaporation'' and pheromone deposits from the ants. This means there is no
guarantee a greedily selected path one iteration will be the same as the
greedily selected path from the previous iteration. In the initial stages of the
algorithm, these paths may be quite different.

In this work, we propose a greedy variant of the Beam-ACO algorithm, referred to
as gBeam-ACO, that replaces the stochastic beam search heuristic of Beam-ACO with
a greedy approach. The primary motivation of this approach is to avoid costly operations
associated with PRNGs. We show that replacing the stochastic heuristic with a
greedy heuristic improves runtime performance by more than 50\% while still
maintaining the quality of solutions. The trade-off with this approach, as with
all greedy heuristics, is the algorithm focuses more on exploitation rather than
exploration. This means the beam-width hyper-parameter of the Beam-ACO algorithm
is important because it gives the algorithm some degree of
exploration. This greedy approach works in the ACO setting because heuristic
weights of the paths are updated during the initial phases of the search.
Thus, although the algorithm is greedy, it is likely
that it will not visit the same paths until the pheromone matrix has stabilized.


\section{Prior Work}
Most similar to our work is that of Beam-ACO
\cite{blum2005beam,lopez2010beam,blum2008beam,caldeira2007supply,thiruvady2009hybridizing}.
Our work
differs from the original by proposing the use of a greedy heuristic during beam
selection rather than using a stochastic heuristic. This results in an improved
runtime, which consequently leads to better found solutions for the same
running time and iteration count.

Application of a greedy heuristic to ACO was explored in
\cite{DBLP:conf/globecom/VendraminMDV11,VENDRAMIN2012997}.
Vendramin et al. propose a greedy variant of the ACO heuristic as a routing
protocol in delay tolerant networks.
They use an event-based mechanism to decay pheromones, which is is an important
difference to our work. In our work we want pheromone levels to decay at each time step
to ensure we don't get stuck in a sub-optimal path. As proposed in \cite{blum2005beam},
using an MMAS-style pheromone update rule means every feasible path remains
feasible because there is always a non-zero pheromone level. Additionally,
our use of beam search means we are less likely to get stuck in a sub-optimal solution
resulting from a greedy path extension since we consider a number of partial
paths simultaneously.

Greedy ACO algorithms have also been studied in the design of wide sensor networks
\cite{DBLP:journals/jnca/LiuH14} and networking \cite{chengming2013greedy}.
In this setting, the idea is to prefer vertices
that have a larger number of neighboring vertices within range of their network.
While this is
technically a greedy heuristic (greedy in the number of reachable neighbors), it
has the effect of increasing exploration rather than increasing exploitation, which
is the usual effect of choosing a greedy algorithm. In this sense, the approach is
very different from our work. We are not choosing vertices based on their number of
neighbors; because we are working on a complete graph this heuristic is not 
effective. Additionally, we focus solely on the pheromone heuristic to guide the search.
Our algorithm focuses on exploring the top-$k$  most promising paths (for beam width $k$)
as defined by the ACO heuristic function.

While there are some works that apply greedy modifications to ACO in the context
of the Traveling Salesman Problem, most work involving greedy algorithms are for
more classical search techniques. For example, \cite{GENG20113680} applies a
greedy selection algorithm with an adaptive simulated annealing
\cite{Kirkpatrick83optimizationby} algorithm.

\section{Beam-ACO}
The combination of beam search with ACO was introduced in \cite{blum2005beam}
and described by Algorithm~\ref{alg:beam-aco}.
ACO iteratively constructs solutions by randomly sampling from the distribution
of heuristic weights, described in Algorithm~\ref{alg:stochastic-partial-aco},
probabilistically guided by heuristic information \cite{dorigo2004ant}. The heuristic
function used by ACO is the weighted
product of the reciprocal of the edge length and the current pheromone level
of the respective edge. This is shown equation~\eqref{eq:heuristic}, where
$\alpha$ and $\beta$ are constants chosen by the user, $P(x, y)$ is the
pheromone level on the edge connecting node $x$ and node $y$, and $E(x, y)$
is the length of the edge between $x$ and $y$.
\begin{equation}\label{eq:heuristic}
  h(x, y) = P(x, y)^{\alpha}E(x, y)^{\beta}
\end{equation}
We use $\alpha = 1$ and $\beta = 4$, based on the analysis from \cite{DBLP:conf/icai/GaertnerC05}
but note that the optimal values of these parameters is highly problem dependent.
Anecdotaly, we found these values of $\alpha$ and $\beta$ produced better results
for all ACO algorithms in our experiments than using $\alpha=1$ and $\beta=1$.
Beam-ACO selects the edge stochastically by normalizing the heuristic weights
for each edge and setting $h(x, y)$ to $0$ if $y$ has already been visited.
The probability distribution used by Beam-ACO for selecting a new vertex
$y$ is given by equation~\eqref{eq:aco-dist}, where
$V$ is the set of vertices of the graph $G=(V, E)$ for the respective TSP
and $P$ is the set of previously visited vertices.
\begin{equation}\label{eq:aco-dist}
  p(y|x) = \begin{cases}
    0 & y \in P\\
    h(x, y) / \sum_{y' \in V \setminus P} h(x, y')  & \text{otherwise}
  \end{cases}
\end{equation}
Note that the current vertex is considered visited, which means there is no chance of
selecting the current vertex.
Beam-ACO
incorporates a probabilistic beam search to expand partial solutions for each
ant. Prior work on beam search
\cite{wilt2010comparison} noted that it is particularly effective for large
problems and we observed this in our experiments. Although both Beam-ACO and
gBeam-ACO found better solutions than both Elitist and MMAS, the gap widened
quite dramatically as the TSP instance size grew.
Define $k_{\text{ext}}$ as the number of extensions (i.e., generated partial solutions)
from the current vertex and define
$k_{\text{bw}}$ as the beam width (i.e., the number of partial solutions maintained between
iterations).
This differs somewhat from classical Beam search in that
we restrict the exploration to $k_{\text{ext}}$ rather than explore the entire
frontier, which consists of all unvisited nodes. This reduces
exploration but also improves runtime efficiency. In our algorithm we set
$k_{\text{ext}} = k_{\text{bw}}$ and refer to the common value as the beam width,
denoted by $k$. Paths are pruned based their current length.

\begin{algorithm}
  \caption{Beam-ACO algorithm.}\label{alg:beam-aco}
  \begin{algorithmic}[1]
    \Procedure{Beam-ACO}{$G, n_{\text{search}}, n_{\text{paths}}, p(s,i)$}
      \State $G$ - TSP graph
      \State $n_{\text{search}}$ -- \# of beam search candidates
      \State $n_{\text{paths}}$ -- \# of maintained paths
      \State $p(s)$ -- predicate that specifies when to stop searching
      \State $\mathcal{S} \gets \emptyset$ -- candidate solutions, $|\mathcal{S}| \leq
      n_{\text{ants}}$
      \State $\texttt{i} \gets 0$\Comment{Current iteration.}
      \State \emph{// Repeat until stopping predicate it true}
      \While{\textbf{not} $p(\mathcal{S}, \texttt{i})$}
        \State $\texttt{i} \gets \texttt{i} + 1$
        \For{$s \in \mathcal{S}$}
          \State \emph{// Extended solution stochastically or greedily}
          \State $\mathcal{S} \gets \mathcal{S} \cup
            \texttt{ExtendAcoSolution}(s, k, G, \texttt{SolutionType})$\label{alg:extend}
        \EndFor
        \State $n \gets \min(n_{\text{paths}}, |\mathcal{S}|)$
        \State $\mathcal{S} \gets \texttt{PickTopN}(\mathcal{S}, n)$
        \State $G.\texttt{pheromones} \gets \texttt{PheromoneUpdateRule}(\mathcal{S}, G)$
        \EndWhile
        \State \textbf{return} $\arg\min_{s\in S}||s||$
      \EndProcedure
      \Procedure{ExtendAcoSolution}{$s, G, \texttt{SolutionType}$}
      \State \texttt{SolutionType} - an enum of either \texttt{STOCHASTIC} or
      \texttt{GREEDY}
      \If{\texttt{SolutionType} is \texttt{STOCHASTIC}}
      \State \textbf{return} \texttt{StochasticExtendAcoSolution}(s, k, G)
      \Else
      \State \textbf{return} \texttt{GreedyExtendAcoSolution}(s, k, G)
      \EndIf
      \EndProcedure
  \end{algorithmic}
\end{algorithm}

\begin{algorithm}
  \caption{Stochastic extension of partial ACO solution.}\label{alg:stochastic-partial-aco}
  \begin{algorithmic}[1]
    \Procedure{StochasticExtendAcoSoltuion}{$s, k, G$}
      \State $s$ -- current partial solution
      \State $G$ -- graph data structure containing distances and pheromone
      \State $G.\texttt{pheromones}$ -- the pheromone matrix
      \State $\mathcal{S} \gets \emptyset$ -- new partial paths
      levels
      \State \emph{// Most recently added node in candidate solution.}
      \State $n \gets s.\texttt{currentNode}$
      \For {$i = 1$ \textbf{t} $k$}
      \State \emph{//$+$ operator is clone and append}
      \State $\mathcal{S} \gets \mathcal{S} \cup s + \texttt{RandomSample}(G.\texttt{pheromones}[n])$\label{alg:sample}
      \EndFor
      \State \textbf{return} $\mathcal{S}$
    \EndProcedure
  \end{algorithmic}
\end{algorithm}

We found Beam-ACO to be an effective ACO algorithm, returning better results
than both Elitist and MMAS ACO algorithms. As discussed in greater detail in
Section~\ref{sec:results}, Beam-ACO does more work per iteration than the
standard ACO algorithms, which means Beam-ACO cannot be compared with these
standard ACO algorithms on a per-iteration basis.

The reason Beam-ACO is so effective with respect to other
ACO algorithms is that the beam search component forces the algorithm to explore
a larger number of partial paths. For a single ant in a typical ACO algorithm, a
path is constructed probabilistically vertex by vertex. A Beam-ACO ant with beam
width of $k$ will select
$k$ vertices for each partial path it is constructing. This
means after the first step it will consider a total of $k^2$
paths and select the $k$ shortest of those paths. Compared
to a colony of $n$ ants running a standard ACO algorithm, a Beam-ACO ant will
explore a much larger variety of partial paths. This is why Beam-ACO is so
effective at finding better solutions than the typical ACO algorithms, but it is
also why we must be careful in our performance measurements. We want to make
sure we compare algorithms when they are doing equivalent amounts of work.

\subsection{Greedy Beam-ACO}
Modifying the Beam-ACO algorithm for a greedy heuristic is very straightforward:
instead of sampling during the path construction section of the algorithm
(Algorithm~\ref{alg:stochastic-partial-aco}, line~\ref{alg:sample}) we
simply select the $k$ paths with the largest heuristic weights for a beam-width
of $k$. The full description of the gBeam-ACO algorithm is the same as that
shown in Algorithm~\ref{alg:beam-aco}, except we replace the
\texttt{ExtendAcoSolution} call with the \texttt{GreedyExtendAcoSolution}
procedure described in Algorithm~\ref{alg:greedy-partial-aco}. This is done
by passing an enum as one of the arguments to \texttt{ExtendAcoSolution},
which can take on the values \texttt{SolutionType.STOCHASTIC} or
\texttt{SolutionType.GREEDY}.

There are a two
motivations for using this greedy approach over a stochastic selection algorithm.
First, we can improve runtime performance by avoiding calls to a PRNG without using
a complex alternative, such as using a buffered channel to communicate random numbers
generated in a separate thread. Secondly, because we are greedily selecting based
on the heuristic edge weight $h(x, y)$, and $h(x, y)$ changes each iteration, we
force selection and evaluation of the $k$ most promising paths for the current iteration.
Maintaining the beam in this setting helps the algorithm overcome some of the exploitation-bias.

There is extensive prior work benchmarking and improving runtime performance and parallelization
of PRNGs \cite{mascagni2000algorithm,de1988parallelization,coddington1997random}.
Despite this progress, skipping the call to the PRNG will always be faster (i.e., no work is
better than some work). There is always a trade-off and in this case it is in the quality of
the solution. As we will demonstrate experimentally, on average gBeam-ACO's solutions are
within 5\% of Beam-ACO's solutions (and occassionally better than Beam-ACO) but it finds
these solutions nearly an order of magnitude faster than Beam-ACO.
We found that avoiding the PRNG alone was worth about an 8-10\% performance boost.

One interesting consequence of using a greedy selection algorithm
when constructing the path is that every ant will construct the same set of partial
paths and return the same final path. Table~\ref{tab:results-ab} shows this
consequence: ``gBeam-10x1'' is gBeam-ACO with a beam width of 10 and 1 ant
while ``gBeam-10x10'' is gBeam-ACO with a beam width of 10 and 10 ants, both
have identical path length numbers.
This means there is no benefit to using more than one
ant, which means that gBeam-ACO, when compared with Beam-ACO using $n$ ants,
should at least see around a $(n-1)/n$
reduction in its runtime simply from only having to maintain a single ant. This consequence
is where gBeam-ACO gets the majority of its runtime performance improvement. It is also
interesting to note that despite only using a single ant with a greedy path construction
heuristic it is able to find similar quality solutions to standard Beam-ACO using multiple
ants.

gBeam-ACO finds high quality solutions using a greedy heuristic in a problem setting
where greedy algorithms typically struggle because it has a guaranteed level of exploration
from maintaining the beam of partial solutions. Additionally, the pheromone levels change
each iteration, leading to new greedily constructed solutions. The changing
pheromone levels make it less likely that a greedily constructed path
resulting in a poor solution will be constructed again because the pheromones on the edges
of the respective path will all decrease before the next iteration. While the changing
pheromone levels help gBeam-ACO explore the graph, maintaining the search beam is the key aspect
of the algorithm that offsets the exploitation-bias of the greedy path construction heuristic.
During partial path construction, a gBeam-ACO ant will explore $k$ new partial paths for each
of its current $k$ partial paths. This means for a beam width of $k=10$, each partial path
extension step will consider 100 new partial paths and then prune them down to the 10 shortest.
A corresponding Beam-ACO ant will do the same exploration, but also make 100 calls to a PRNG.
Additionally, Beam-ACO will repeat this 10 times generating 1,000 new and unique partial paths and pruning
them down to 100 (10 partial paths per ant). Given the disparity in work between the two
algorithms, it is somewhat surprising that gBeam-ACO is able to find solutions of similar
quality.

\begin{algorithm}
	\caption{Greedy extension of partial ACO solution.}\label{alg:greedy-partial-aco}
	\begin{algorithmic}[1]
		\Procedure{GreedyExtendAcoSolution}{$s, k, G$}
	    \State $s$ -- current partial solution
        \State $k$ -- beam width
      \State $G$ -- Data structure storing TSP and ACO metadata
      \State $S \gets \emptyset$ \Comment{Empty set to store partial paths}
      \State $n \gets s.\texttt{currentNode}$ \Comment{Most recently added node
      in candidate solution.}
			\State $\texttt{minNode} \gets \texttt{FindMaxNotInSet}(n.\texttt{neighbors}, -\infty$)
			\For{$i = 1$ \textbf{to} $k$}
				\State $\texttt{minNode} \gets
                \texttt{FindMaxNotInSet}(n.\texttt{neighbors},
                \texttt{minNode}$)
				\State $S \gets S \cup s.\texttt{cloneAndAppend(minNode)}$
			\EndFor
      \State \textbf{return} $S$
		\EndProcedure
	\end{algorithmic}
\end{algorithm}

\begin{algorithm}
	\caption{Max-finding algorithm: finds $\max xs\setminus ys$.}\label{alg:find-min}
	\begin{algorithmic}[1]
		\Procedure{FindMaxNotInSet}{$xs, ys$}
			\State $xs$ - collection of TSP node indices
			\State $ys$ - set of previously selected indices
			\State $x' \gets xs[0]$ \Comment{Set to initial element}
			\State $w' \gets \texttt{HeuristicWeight}(x')$
			\For{$x \in xs$}
				\State $w' \gets \texttt{HeuristicWeight}(x)$
				\If{$w' > w \textbf{ and } x' \not\in ys$}
					\State $(x', w') \gets (x, w)$
				\EndIf
			\EndFor
			\State $\textbf{return } x'$
		\EndProcedure
	\end{algorithmic}
\end{algorithm}


\subsection{Complexity}\label{sec:complexity}
Suppose we have $n$ ants, a beam width of $k$, and a TSP with $|V|$
vertices. The ACO algorithm will consider $O(n|V|^2)$ partial paths
per iteration while the Beam-ACO algorithm will consider $O(n|V|^2k^2)$
partial paths. The gBeam-ACO iteration complexity is slightly different from Beam-ACO
because it only needs a single ant and its greedy update rule, which takes
$O(|V|k)$ time.
Thus the gBeam-ACO complexity is given by $O(|V|^2k^2)$.
\begin{figure*}[t]
	\centering
	\includegraphics[width=.95\textwidth]{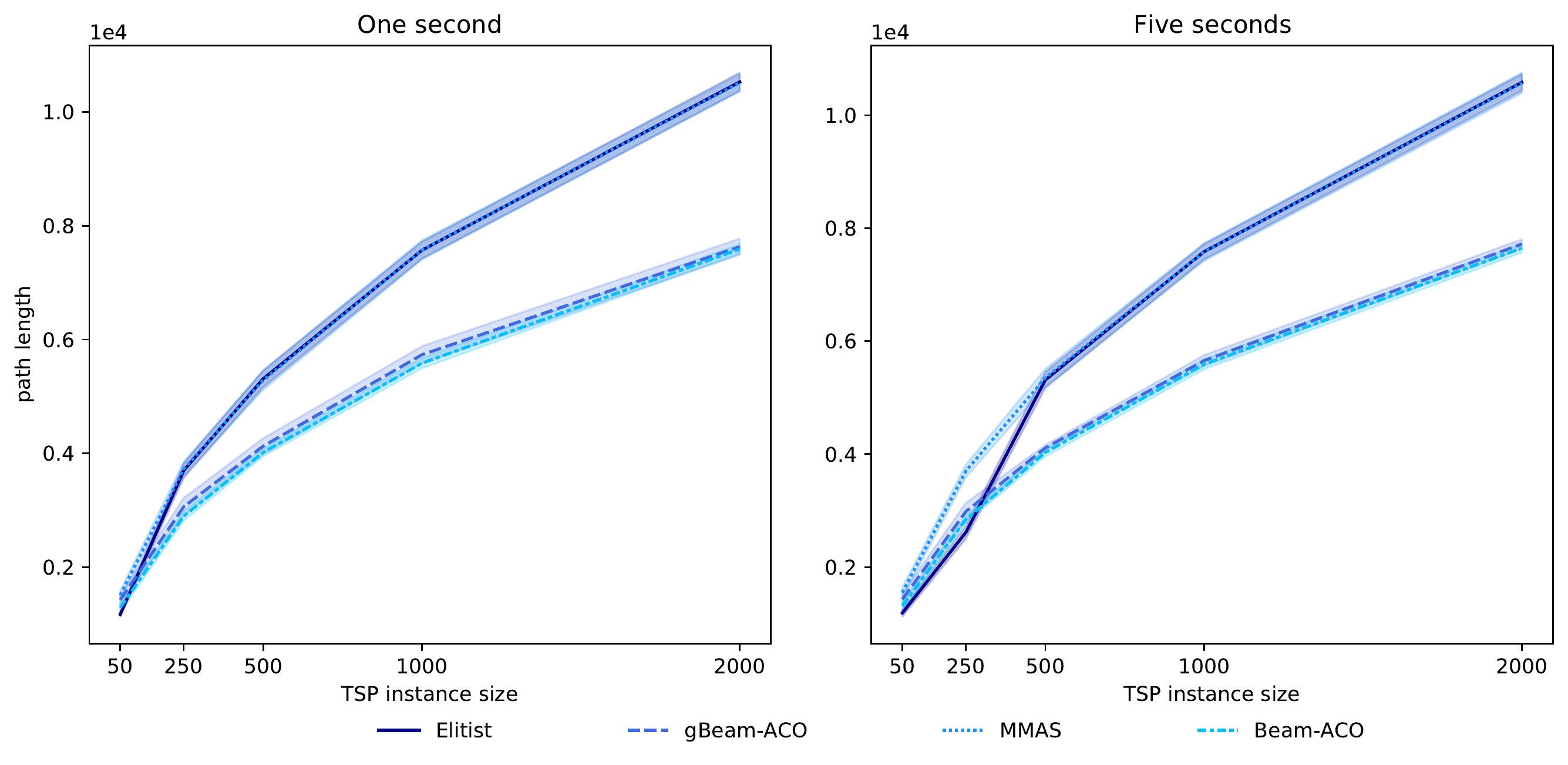}
	\caption{Comparison of ACO algorithms for varying durations.}
	\label{fig:aco-length-comp}
\end{figure*}

\subsection{Trade-offs}
The major trade-off is exploration -- gBeam-ACO is an exploitation-heavy
algorithm, which means it will favor paths it knows are promising over exploring
new paths that may be less promising or do not have much heuristic information. As
previously discussed, because the heuristic weights change each iteration
and the algorithm maintains a beam of partial paths, there is a level of forced
exploration.

The other issue with the exploitation versus exploration trade-off is that once
the pheromone distribution has converged to a stable distribution after some number
of iterations, gBeam-ACO will always produce the same set of solutions--which
may or may not be optimal--while Beam-ACO is capable of producing new,
previously unexplored solutions. We note that both Beam-ACO and gBeam-ACO are
computationally intense ACO algorithms, both of which do substantially more work
per iteration than vanilla ACO algorithms. In our experience, both gBeam-ACO and
Beam-ACO are about $30-50\times$ slower on a per iteration basis for TSP
instances with 500 to 1,000 vertices. In other words, the gBeam-ACO and
Beam-ACO algorithms were only able to complete one or two iterations in the same
time vanilla ACO algorithms such as Elitist Ant System
\cite{DBLP:journals/corr/abs-0811-0131,DBLP:conf/fgit/JaradatA10}
and Max-Min Ant System (MMAS) \cite{10.5555/348599.348603}
 were able to complete 30 to
50 iterations. Despite the difference in iteration count, gBeam-ACO and Beam-ACO
produced dramatically better solutions. The large amount of per-iteration work
performed by gBeam-ACO means it will take a relatively long time (in terms of wall
clock time) for gBeam-ACO's pheromone distribution to converge to a fixed
distribution. It is important to note that
gBeam-ACO seems to be very effective even when it is only run for one or two
iterations.

\section{Results}\label{sec:results}
In our initial work it became apparent that gBeam-ACO was able to achieve
similar results to standard Beam-ACO but in a much shorter time.
Care must be taken in these measurements because these
algorithms are doing a different amount of work per iteration. As an example,
comparing ACO with Beam-ACO on a per-iteration basis is not a fair comparison
because they consider different numbers of partial paths per iteration.
Similarly, we must be careful in measuring the
differences between Beam-ACO and gBeam-ACO because gBeam-ACO only uses a single
ant, which means at minimum it will do $n - 1$ fewer iterations
than Beam-ACO. This makes a fair comparison rather difficult because gBeam-ACO
is doing less work for a fixed iteration count (because there are fewer ants)
but because it has a shorter wall-clock runtime when compared to Beam-ACO it is
able to do more iterations per fixed unit of time.

All experiments use the same hyper-parameter tunings. We set $\alpha = 1$ and
$\beta = 4$. Pheromones evaporate at a rate of $0.1$ and the pheromone deposit
amount is $1.0$. MMAS min and max pheromone levels are 0.1 and 0.9,
respectively. Experiments that use randomly generated TSP instances use the
domain $[-100, 100]\times[-100, 100]$.

Figure~\ref{fig:aco-length-comp} compares Elitist, MMAS, Beam-ACO, and gBeam-ACO
on a fixed duration basis. Fifteen TSP instances were randomly generated
and evaluated by each of the algorithms.
All algorithms ran for approximately one and five seconds. We did not stop
algorithms mid-iteration. This was not an issue for TSP instances 500 vertices
and smaller, but for the 1,000 and 2,000 vertex instances both gBeam-ACO and
Beam-ACO ran longer than the allotted time. In both of these cases gBeam-ACO
and Beam-ACO
only completed a single iteration while Elitist and MMAS both completed
on the order of tens of iterations. This gives some insight as to the value of
maintaining the search beam -- despite both gBeam-ACO and Beam-ACO not being
able to take advantage of the updated pheromone matrix, both were able to find
better solutions than Elitist and MMAS algorithms. Both gBeam-ACO and Beam-ACO
outperform Elitist and MMAS algorithms on larger problem size. Because the
main difference in the algorithms is the beam search component, this matches
previous observations on beam search \cite{wilt2010comparison} that it performs
particularly well on large problem sizes. Although it is interesting that Elitist
finds the best solutions for the 50 and 250 size problems while running for five
seconds, we don't find this particularly damaging to the Beam-ACO family of
algorithms because these are relatively small TSP instances.

\begin{figure}[t]
	\centering
	\includegraphics[width=.75\columnwidth]{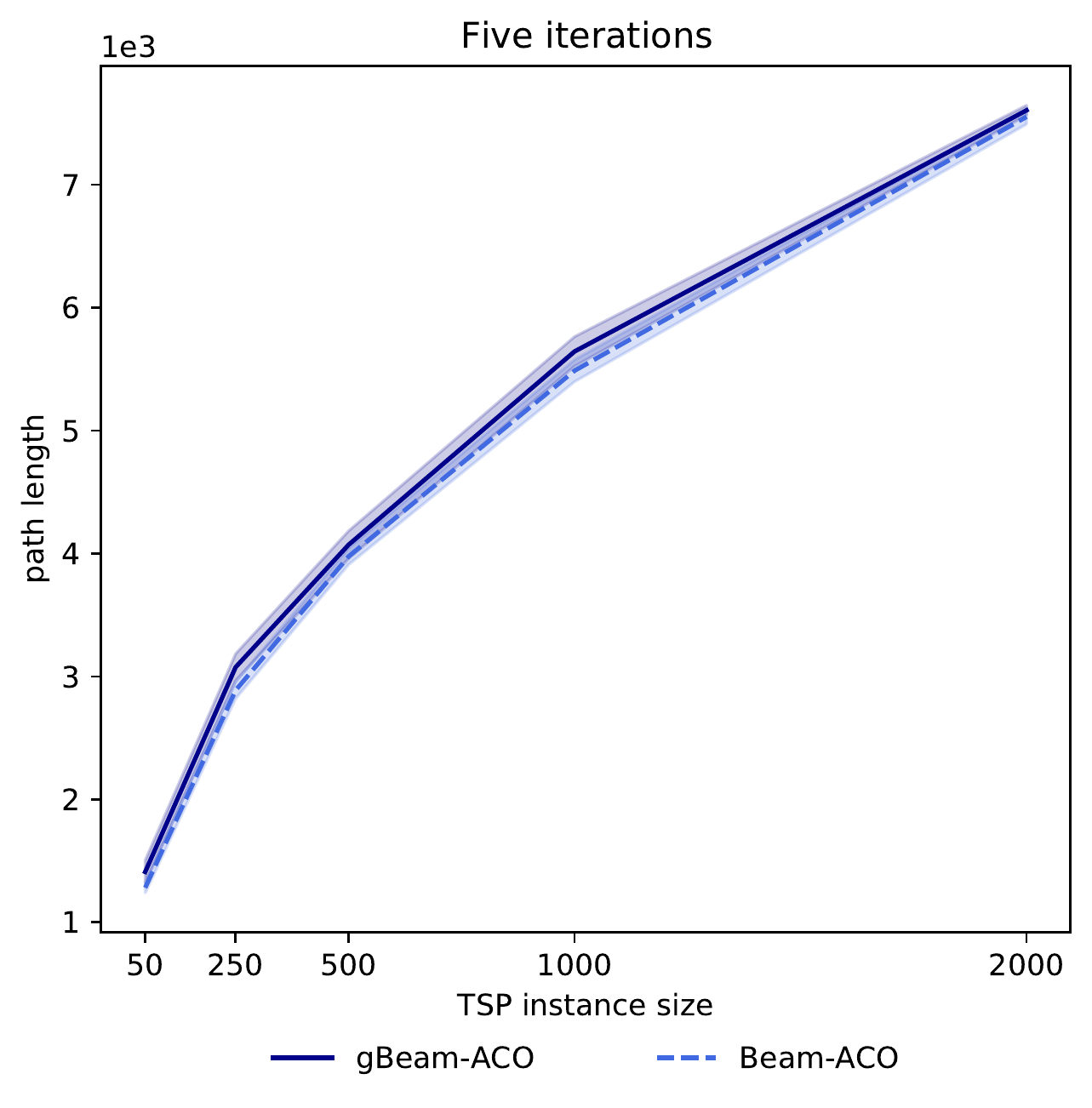}
	\caption{Comparison of best found path length for the same number of iterations.}
	\label{fig:iteration-comp}
\end{figure}

Figure~\ref{fig:iteration-comp} shows the results of the fixed iteration
experiment. This experiment compares Beam-ACO and gBeam-ACO on a per iteration
basis. We do not track runtime because gBeam-ACO has a major advantage. Rather,
we are interested in how the quality of their solutions differ given that one
algorithm is doing more work than the other. These experiments are performed on
15 randomly generated TSP instances.
Two aspects of
Figure~\ref{fig:iteration-comp} stand out: gBeam-ACO finds solutions that are
comparable (although not better) to Beam-ACO and as the problem size grows this
difference becomes smaller. This matches intuition --- Beam-ACO performs more
exploration than gBeam-ACO, so we would expect it to consider a more diverse
population of paths (that is, paths that may not appear as promising based
purely on their heuristic weigh). This diversity means it may find paths that
gBeam-ACO would never consider because of its greedy selection heuristic.

To address the total work disparity, we perform another set of experiments
where a third gBeam-ACO instance whose beam width, $k_{\text{equiv}}$ is set using
equation~\eqref{eq:beam-width}, where $n$ and $k$ are taken from a Beam-ACO instance.
\begin{equation}\label{eq:beam-width}
	k_{\text{equiv}} = \frac{\sqrt{n}}{k}
\end{equation}
This gBeam-ACO instance attempts to do a similar amount of work as a given Beam-ACO instance.
This work largely comes from the partial path extensions performed when
expanding the beam.
For example, if Beam-ACO has 10 ants using a beam width
of 10, then each ant is exploring 100 partial paths, or 1,000 partial paths in
total. Using equation~\eqref{eq:beam-width} yields a beam width of about 32, which
leads to 1,024 partial paths explored at each step by a single ant. Thus,
gBeam-ACO with a beam width of 32 should perform roughly the same amount of work
as Beam-ACO with 10 ants and a beam width of 10.

Table~\ref{tab:results-ab} shows
the results of this experiment, and also includes a gBeam-ACO instance with 10
ants with a beam width of 10. This last instance was included to confirm the
claim that running gBeam-ACO with more than one ant is unnecessary. These
results are based on 10 randomly generated TSP instances. Each algorithm in this
experiment performs five iterations.

The most notable result in this table is that gBeam-ACO with a single ant
is over 90\% faster than Beam-ACO while finding paths that are, on average,
less than 5\% longer. In other words, gBeam-ACO trades a 5\% decrease in quality
of solution for a 90\% increase in speed. Comparing gBeam-ACO with 10 ants to Beam-ACO
shows over a 10\% reduction in runtime almost across the board. This runtime
savings is purely from avoiding calls to the PRNG. This comes at a cost ---
Beam-ACO is able to consistently find shorter paths than gBeam-ACO due to
gBeam-ACO's bias towards exploitation over exploration.

Another interesting and somewhat surprising result is the performance of
gBeam-ACO with a beam width of 32 and a single ant. As expected from equation~
\eqref{eq:beam-width}, the runtime is similar to Beam-ACO. Our expectation was
that the wider beam would result in better solutions when compared to a smaller
beam width gBeam-ACO algorithm and possibly better solutions than standard Beam-ACO.
Although the solutions of the $k = 32$ beam
width are better than $k = 10$, we were surprised that they were not that much
better, especially given the runtime differences. Perhaps what is more
surprising is that the wider beam width gBeam-ACO was not able to consistently
find better solutions than Beam-ACO. This illustrates the importance of
exploration in finding better solutions.

\begin{table*}[]
\centering
\caption{Comparison of several variants of gBeam-ACO with Beam-ACO based on 10 trials of randomly generated TSP instances.}
\label{tab:results-ab}
\begin{adjustbox}{width=\textwidth}
\begin{tabular}{l|cccc|cccc}\toprule
                  & \multicolumn{4}{c|}{\textbf{Path Length}}                                               & \multicolumn{4}{c}{\textbf{Runtime}}\\\midrule
\textbf{TSP Size} & \textbf{Beam-10x10} & \textbf{gBeam-10x1} & \textbf{gBeam-32x1} & \textbf{gBeam-10x10} & \textbf{Beam-10x10} & \textbf{gBeam-10x1} & \textbf{gBeam-32x1} & \textbf{gBeam-10x10} \\\midrule
50                & $\mathbf{1,449 \pm 98}$      & $1,513 \pm 107$     & $1,446 \pm 69$      & $1,513 \pm 107$      & $0.69 \pm 0.02$     & $\mathbf{0.07 \pm 0.00}$     & $0.66 \pm 0.01$     & $0.72 \pm 0.01$       \\
250               & $\mathbf{2,859 \pm 98}$      & $3,052 \pm 97$      & $3,082 \pm 156$     & $3,052 \pm 97$       & $16.63 \pm 0.27$    & $\mathbf{1.54 \pm 0.03}$     & $16.45 \pm 0.20$    & $15.42 \pm 0.36$      \\
500               & $\mathbf{3,961 \pm 87}$      & $4,131 \pm 89$      & $4,129 \pm 129$     & $4,131 \pm 89$       & $68.31 \pm 1.70$    & $\mathbf{6.14 \pm 0.19}$     & $65.07 \pm 1.64$    & $61.29 \pm 1.83$      \\
1,000             & $\mathbf{5,528 \pm 87}$      & $5,670 \pm 80$      & $5,639 \pm 101$     & $5,670 \pm 80$       & $262.41 \pm 5.73$   & $\mathbf{22.74 \pm 0.43}$    & $240.25 \pm 4.96$   & $229.61 \pm 5.84$    \\\bottomrule
\end{tabular}
\end{adjustbox}
\end{table*}

Table~\ref{tab:tsps} compares Elitist and MMAS ACO algorithms against Beam-ACO
and gBeam-ACO on 10 TSP instances taken from TSPLIB.
Again, comparison between these algorithms is challenging because
they perform different amounts of work per iteration. Beam-ACO has the largest
per-iteration runtime cost, so we structured the runs of Elitist and MMAS ACO
algorithms such that they would run for a similar wall-clock time as Beam-ACO.
We restricted gBeam-ACO to either one second of runtime or a single iteration,
whichever takes longer. This puts gBeam-ACO at a disadvantage in terms of
quality of solution because gBeam-ACO is running for a tenth of the time of
the other algorithms (excluding the first three instances), but as the results show gBeam-ACO still performs
surprisingly well given these restrictions.

Of the ten TSP instances, gBeam-ACO found the best solution in 30\% of the
time, Beam-ACO found the best solution 50\% of the time, and Elitist ACO found
the best solution 20\% of the time. Across the board, gBeam-ACO consistently had
runtimes that were over 90\% shorter than Beam-ACO's runtime due to only using
a single ant and avoiding PRNG calls. Recall that gBeam-ACO was restricted to
run for either a second or a single iteration depending on which takes
longer. If the runtime is over a few seconds that indicates that gBeam-ACO took
longer than a second to complete a single iteration. Because gBeam-ACO has a
dramatically shorter per-iteration runtime compared to Beam-ACO, we can infer
that Beam-ACO only completed a single iteration in these settings as well. For
these problems--every problem except a280, ch130, and ch150--gBeam-ACO was
nearly 10 times faster than Beam-ACO. For these same
problems, gBeam-ACO found the best solution three out of seven times and was
within about 1\% of the best found solution (found by Beam-ACO) for three of the
remaining four problems (fnl4461, rl1323, rl5915, u2319).

\begin{table*}[]
\centering
\caption{Results from TSP instances selected from TSPlib.}
\label{tab:tsps}
\begin{adjustbox}{width=\textwidth}
\begin{tabular}{l|cccc|cccc|cccc}\toprule
                      & \multicolumn{4}{c|}{\textbf{Path Length}}                          & \multicolumn{4}{c|}{\textbf{Runtime (s)}}                          & \multicolumn{4}{c}{\textbf{1,000 Partial Paths / s}}                    \\\midrule
\textbf{TSP Instance} & \textbf{Elitist} & \textbf{MMAS} & \textbf{Beam} & \textbf{gBeam} & \textbf{Elitist} & \textbf{MMAS} & \textbf{Beam} & \textbf{gBeam} & \textbf{Elitist} & \textbf{MMAS} & \textbf{Beam} & \textbf{gBeam} \\\midrule
a280                  & 3,276             & 4,327          & \textbf{3,254}          & 3,399           & 5.0           & 5.0        & 4.3        & \textbf{1.2}         & \textbf{75}        & \textbf{75}     & 64     & 65      \\
ch130                 & \textbf{6,671}             & 9,149          & 7,505          & 7,712           & 2.0           & 2.0        & 1.9        & \textbf{1.0}         & \textbf{159}       & 158    & 131    & 132     \\
ch150                 & \textbf{7,748}             & 10,956         & 8,419          & 8,450           & 2.0           & 2.0        & 1.3        & \textbf{1.0}         & 136       & \textbf{137}    & 114    & 117     \\
d1291                 & 84,995            & 92,803         & 60,859         & \textbf{59,434}          & 88.6          & 88.0       & 87.6       & \textbf{7.9}         & \textbf{16}        & \textbf{16}     & 14     & \textbf{16}      \\
d1655                 & 99,258            & 115,400        & 78,804         & \textbf{78,783}          & 146.4         & 146.9      & 145.8      & \textbf{12.9}        & \textbf{12}        & \textbf{12}     & 11     & \textbf{12}      \\
fnl4461               & 313,311           & 348,401        & \textbf{237,062}        & 241,913         & 1,087.5          & 1,093.2       & 1,084.1       & \textbf{92.1}        & 4         & 4      & 4      & \textbf{5}       \\
rl1304                & 474,377           & 501,293        & \textbf{336,368}        & 357,146         & 87.2          & 87.4       & 86.4       & \textbf{7.1}         & 17        & 17     & 15     & \textbf{18}      \\
rl1323                & 497,934           & 525,898        & \textbf{360,502}        & 362,975         & 89.7          & 89.3       & 88.6       & \textbf{7.2}         & 16        & 16     & 15     & \textbf{18}      \\
rl5915                & 1,140,250          & 1,183,545       & \textbf{754,365}        & 765,015         & 1,883.9        & 1,888.3     & 1,876.2     & \textbf{143.8}       & 3         & 3      & 3      & \textbf{4}       \\
u2319                 & 396,029           & 420,631        & 292,107        & \textbf{291,531}         & 275.3         & 276.1      & 273.1      & \textbf{23.0}         & 9         & 9      & 8      & 10     \\\bottomrule
\end{tabular}
\end{adjustbox}
\end{table*}

\section{Conclusion}
We have introduced a greedy alternative to Beam-ACO and have
demonstrated its effectiveness in a number of settings. gBeam-ACO is
able to achieve comparable results (generally within 5\% of Beam-ACO in our experiments)
in a fraction of the time. There are two avenues of future work based on these
promising results: (1) exploring a hybrid algorithm that uses gBeam-ACO to quickly
find good solutions and uses standard Beam-ACO to fully explore
those solutions and (2) exploring an asynchronous gBeam-ACO implementation where
ants continuously share pheromone updates, which would allow gBeam-ACO to take
advantage of larger colony sizes. gBeam-ACO's performance make it a great option
for exploratory analysis or bootstrapping pheromone matrices for other algorithms.

\bibliographystyle{unsrt}
\bibliography{bibliography}

\end{document}